\documentclass[]{spie}  %>>> use for US letter paper
\usepackage{amsmath,amsfonts,amssymb}
\usepackage{graphicx}
\usepackage{hyperref}

\title{Non-Learning based Deep Parallel MRI Reconstruction (NLDpMRI)}

\author[*1,2]{Ali Pour Yazdanpanah}
\author[1,2]{Onur Afacan}
\author[1,2]{Simon K. Warfield}
\affil[1]{Computational Radiology Laboratory, Boston Children's Hospital, Boston, MA, 02115, USA.}
\affil[2]{Harvard Medical School, Boston, MA, 02115, USA.}

\authorinfo{Further author information: (Send correspondence to Ali Pour Yazdanpanah)\\ *Ali Pour Yazdanpanah: E-mail: Ali.PouryazdanpanahKermani@childrens.harvard.edu\\}

\begin{document} 
\maketitle

\begin{abstract}
Fast data acquisition in Magnetic Resonance Imaging (MRI) is vastly in demand and scan time directly depends on the number of acquired $k$-space samples. Recently, the deep learning-based MRI reconstruction techniques were suggested to accelerate MR image acquisition. The most common issues in any deep learning-based MRI reconstruction approaches are generalizability and transferability. For different MRI scanner configurations using these approaches, the network must be trained from scratch every time with new training dataset, acquired under new configurations, to be able to provide good reconstruction performance. Here, we propose a new generalized parallel imaging method based on deep neural networks called NLDpMRI to reduce any structured aliasing ambiguities related to the different $k$-space undersampling patterns for accelerated data acquisition. Two loss functions including non-regularized and regularized are proposed for parallel MRI reconstruction using deep network optimization and we reconstruct MR images by optimizing the proposed loss functions over the network parameters. Unlike any deep learning-based MRI reconstruction approaches, our method doesn't include any training step that the network learns from a large number of training samples and it only needs the single undersampled multi-coil $k$-space data for reconstruction. Also, the proposed method can handle $k$-space data with different undersampling patterns, and the different number of coils. Experimental results show that the proposed method outperforms the current state-of-the-art GRAPPA method and the deep learning-based variational network method.  
\end{abstract}

% Include a list of keywords after the abstract 

\section{INTRODUCTION}
\label{sec:intro}  % \label{} allows reference to this section

Long scan time is a primary disadvantage of Magnetic Resonance Imaging (MRI). Parallel imaging (PI) techniques have become popular strategies for reducing MRI scan time. Two conventional PI reconstruction algorithms are the SENSE [\citenum{pruessmann1999sense}] and the generalized autocalibrating partially parallel acquisitions (GRAPPA) [\citenum{griswold2002generalized}]. On the other hand, compressed sensing (CS)-based methods [\citenum{lustig2007sparse, liang2009accelerating, ramani2011parallel, yazdanpanah2017compressed, yazdanpanah2017compressed2}] seek to exploit intrinsic image properties of sparsity in a transform domain and have allowed for accelerated imaging in some settings. These techniques can be formulated similarly to a regularized SENSE reconstruction, but use different regularization strategies to increase data acquisition speed while generating better reconstructions. However, recent data-driven methods based on deep learning has resulted in promising improvements in image reconstruction algorithms.

Two primary deep neural networks-based MRI reconstruction frameworks include image-domain-based [\citenum{wang2016accelerating,schlemper2018deep,hammernik2018learning}] and $k$-space-based [\citenum{cheng2018deepspirit}] frameworks. In [\citenum{wang2016accelerating}], the authors used the convolutional neural network (CNN) either as an initialization or a regularization term for constrained reconstruction. In [\citenum{schlemper2018deep}], a cascade of CNNs is presented and trained, and the reconstruction process is considered as a de-aliasing problem in the image domain. In [\citenum{hammernik2018learning}], a variational network reconstruction method is presented and trained for accelerated multi-coil MRI data. In [\citenum{cheng2018deepspirit}], a deep network is applied entirely in the $k$-space domain to utilize known properties in different ways to compensate for missing $k$-space data. On the other hand, the method presented in [\citenum{zhu2018image}] exploited the combination of fully connected layers plus convolutional autoencoder in order to find the direct mapping from the $k$-space domain into the image domain and trained their network with data that modulated with synthesized-phase.
In all the mentioned papers, the deep network needs to learn from new massive training datasets acquired under new configurations through the training process every time from scratch in order to be able to reconstruct efficiently. 

Not being able to generalize to new datasets is the main disadvantage of these type of methods which make them unsuitable in practice especially since a wide range of MR system and protocols exists. The deep learning-based reconstruction methods are sensitive to any deviation between training and test datasets. Especially an SNR deviation between training and test datasets will leads to a considerable reduction of image quality [\citenum{knoll2019assessment}]. Also, any changes in $k$-space sampling pattern will result in an immediate failure for any learning-based reconstruction method. Since having different acquisition parameters in MRI systems are very common between different institutions, the learning-based approaches are not considered as practical solutions for MRI reconstruction.

Here, we propose a new generalized parallel imaging method based on deep neural networks without using any training datasets. Unlike most deep learning-based MRI reconstruction methods, our method operates on real-world acquisitions with the complex data format, not on simulated data, real-valued data, or data with added simulated-phase. We categorize our method among the unsupervised energy-based methods [\citenum{golts2018deep,ulyanov2017deep}]. Using our proposed method, we evaluate the reconstruction performance compared to clinically-used GRAPPA reconstruction method and the recently published deep learning-based variational network (VN) method. 

\section{METHODS}

\begin{figure} [ht]
\begin{center}
\begin{tabular}{c} %% tabular useful for creating an array of images 
\includegraphics[height=9.5cm]{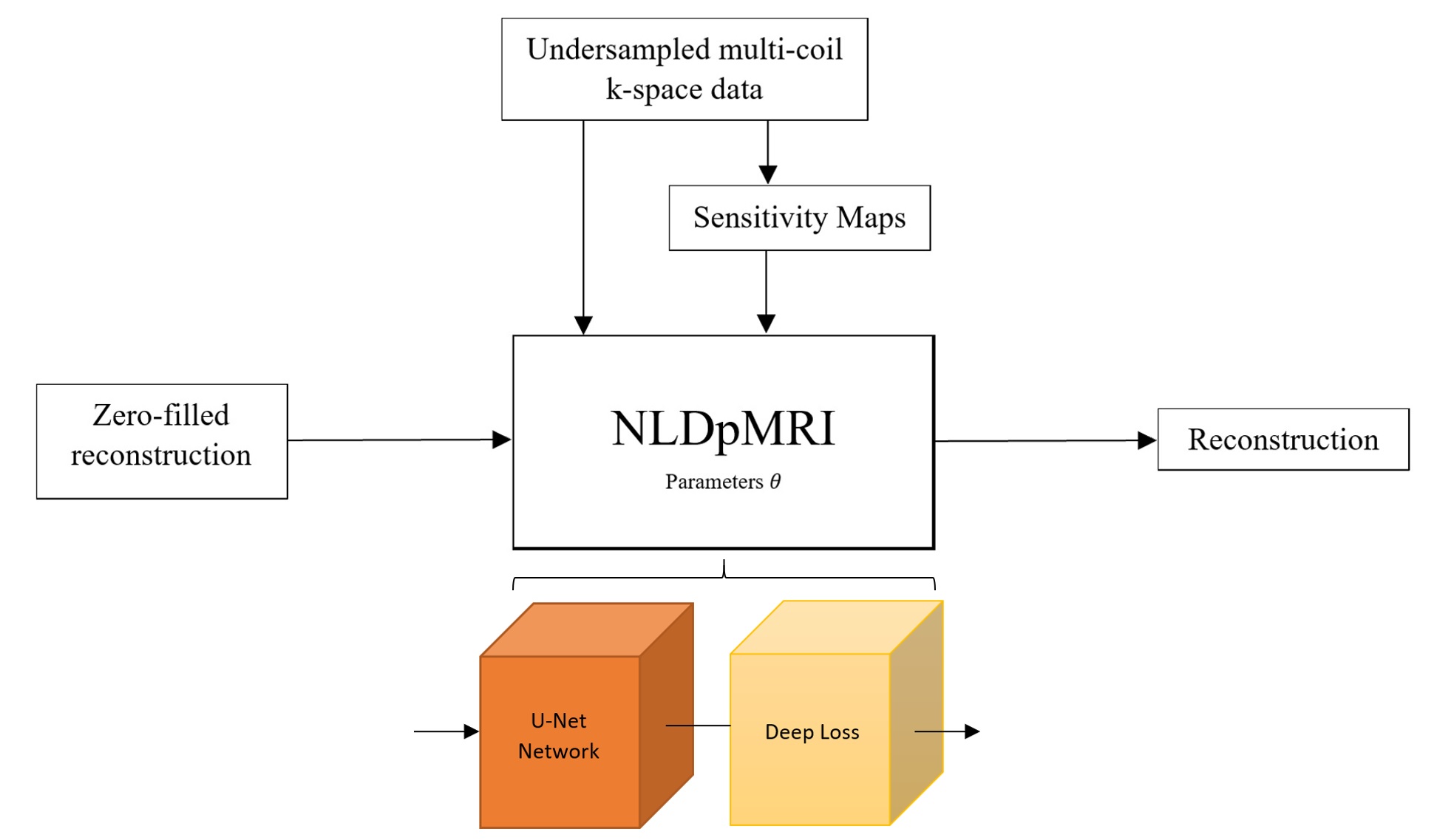}
\end{tabular}
\end{center}
\caption[fig1] 
{ \label{fig1} 
The NLDpMRI reconstruction scheme.}
\end{figure}

We develop the deep neural network-based method, without any training data involved, using encoder-decoder U-net [\citenum{ronneberger2015u}] convolutional network architecture with skip connections for parallel MRI reconstruction (Figure \ref{fig1}). The number of filters for both encoder and decoder layers is set to 128 and the network filter kernel size is set to 3 for both encoder and decoder layers. Only undersampled multi-coil $k$-space raw data is needed for reconstruction. We initialized U-net network parameters randomly and zero-filled reconstruction is used as the network input. Since the deep network frameworks work on real-valued parameters, inputs, and outputs, in our method complex $k$-space data are divided into real and imaginary parts and considered as two-channel input and output. 

Two deep loss functions are proposed based on MRI imaging model. Given the deep loss function, we optimize the deep loss over the network parameters at run-time per subject. The following equations are the proposed non-regularized (Eq.~1) and regularized (Eq.~2) loss functions:

\begin{equation} \label{Eq1}
\begin{aligned}
\hat{\theta} =\underset{\theta}{ arg \text{min}} \ 
\|\text P\text F\text S_{l}\text x_{\theta}-\text d_u\|_{2}^{2},
\end{aligned}
\end{equation}

\begin{equation} \label{Eq2}
\begin{aligned}
\hat{\theta} =\underset{\theta}{ arg \text{min}} \ 
\|\text P\text F\text S_{l}\text x_{\theta}-\text d_u\|_{2}^{2}+ \lambda\|\theta\|_{2}^{2}
\end{aligned}
\end{equation}

where $\text d_u$ is the undersampled $k$-space data. $\theta$ represents the network parameters, $\text x_{\theta}$ is the network output, and $\text x_{\hat{\theta}}$  is the MR image to be reconstructed. $\text P$ is a mask representing $k$-space undersampling pattern and $\text F$ is the Fourier transform operator. $\text S_{l}$  represents coil sensitivity maps and $l$ is the total number of coils. The sensitivity maps were computed using ESPIRiT [\citenum{uecker2014espirit}] method applied to only calibration data. 

The reconstruction is an iterative process which minimizes our loss function over the network parameters at run-time per subject. In our reconstruction process, our network parameters are getting updated in every iteration, not through some training step based on training dataset. They are getting updated based on the single multi-coil undersampled $k$-space data by optimizing the loss function over the network parameters. Loss minimization was performed using ADAM [\citenum{kingma2014adam}] optimizer with an update rate of 0.001. The output of the network at each iteration is the reconstructed image at that step and the output will be updated iteratively through the loss function minimization.

\section{EXPERIMENTS}

For testing our method, we used MRI reconstructions for 3D MPRAGE dataset with 32-channel head coil using NLDpMRI reconstruction and compared our results to standard GRAPPA reconstruction. We used a fully sampled MPRAGE data and retrospectively undersampled in both phase encoding dimensions with an acceleration factor of 2x2. Full $k$-space data reconstructed with the adaptive combine method [\citenum{walsh2000adaptive}], was used as our gold standard for comparison.
Also, in order to demonstrate the generalization capability of our method, we have used the entire test datasets (knee datasets) of recently published variational network reconstruction method [\citenum{hammernik2018learning}] and compared our results to their results in terms of the structural similarity index (SSIM) and normalized root-mean-square error (NRMSE). 
The dataset includes 50 test cases from five different sequences including coronal proton-density (PD), coronal fat-saturated (FS) PD, axial FS T2, sagittal FS T2, and sagittal PD. For more detail, regarding the sequence parameters, you can refer to [\citenum{hammernik2018learning}]. The datasets include fully sampled data which then retrospectively undersampled with an acceleration factor of 4.

The advantage of this experiment is to demonstrate that NLDpMRI can easily handle and reconstruct all five test datasets, which include five different sequences with different parameters, without any training datasets and any training steps involved. On the other hand, for VN method five different networks should be trained individually, a one trained network for each specific sequence, using five massive training datasets (one training dataset per each sequence) beforehand so that VN method can reconstruct the test datasets.

\section{RESULTS}
\label{sec:sections}

\begin{figure} [ht]
\begin{center}
\begin{tabular}{c} %% tabular useful for creating an array of images 
\includegraphics[height=5.5cm]{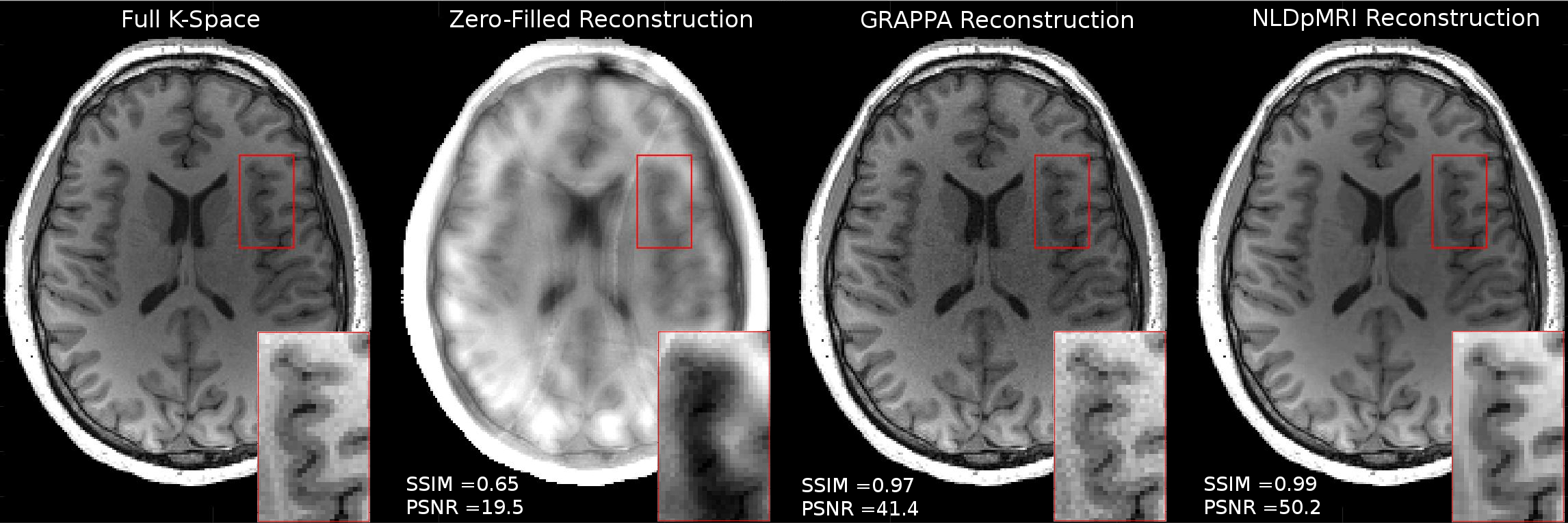}
\end{tabular}
\end{center}
\caption[fig2] 
{ \label{fig2} 
Left to right: Gold standard reconstruction result using full k-space data, zero-filled reconstruction result, GRAPPA reconstruction result, and non-regularized NLDpMRI reconstruction result all with undersampling factor of 2x2. NLDpMRI results in better quality image compared to GRAPPA.}
\end{figure}

\begin{figure} [ht]
\begin{center}
\begin{tabular}{c} %% tabular useful for creating an array of images 
\includegraphics[height=6.5cm]{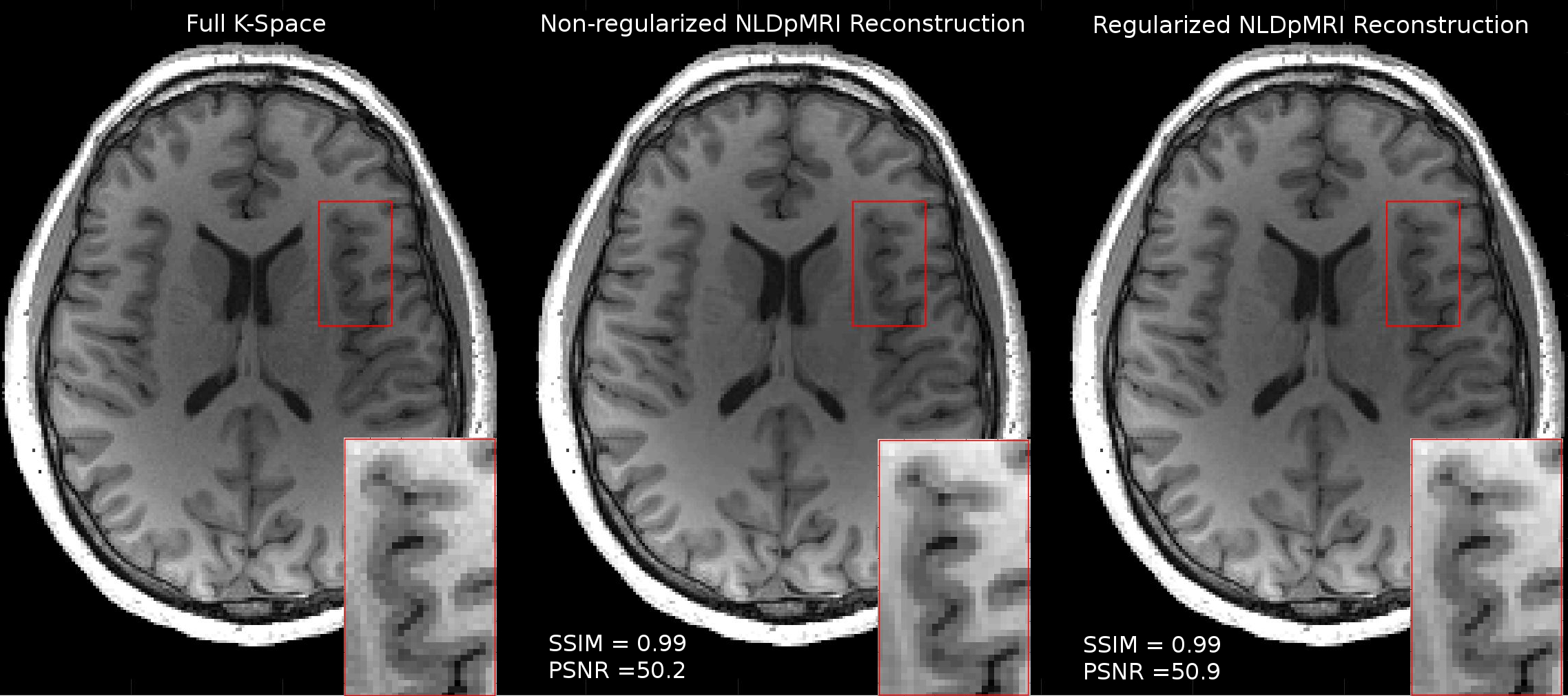}
\end{tabular}
\end{center}
\caption[fig3] 
{ \label{fig3} 
Left to right: Gold standard reconstruction result using full k-space data, non-regularized NLDpMRI reconstruction result, and regularized NLDpMRI reconstruction result all with undersampling factor of 2x2}
\end{figure}

Figure \ref{fig2} denotes the results from our network using the non-regularized loss function and compares the result of our network to the ground truth, zero-filled reconstruction, and GRAPPA reconstruction. Figure \ref{fig3} shows the reconstruction comparison between regularized and non-regularized NLDpMRI reconstruction. We observed that NLDpMRI reconstructs artifact-free images, which have better quality than GRAPPA reconstruction, and GRAPPA result shows noise amplification compared to NLDpMRI result. Additionally, we observed that adding the regularization term to the loss function will slightly improve the reconstruction accuracy (PSNR of regularized NLDpMRI is 50.9 compared to PSNR of 50.2 for non-regularized NLDpMRI). The proposed method uses sensitivities estimated from exactly the same calibration data as GRAPPA method uses (24 x 24 x 24 Cartesian grid). 

\begin{figure} [ht]
\begin{center}
\begin{tabular}{c} %% tabular useful for creating an array of images 
\includegraphics[height=9cm]{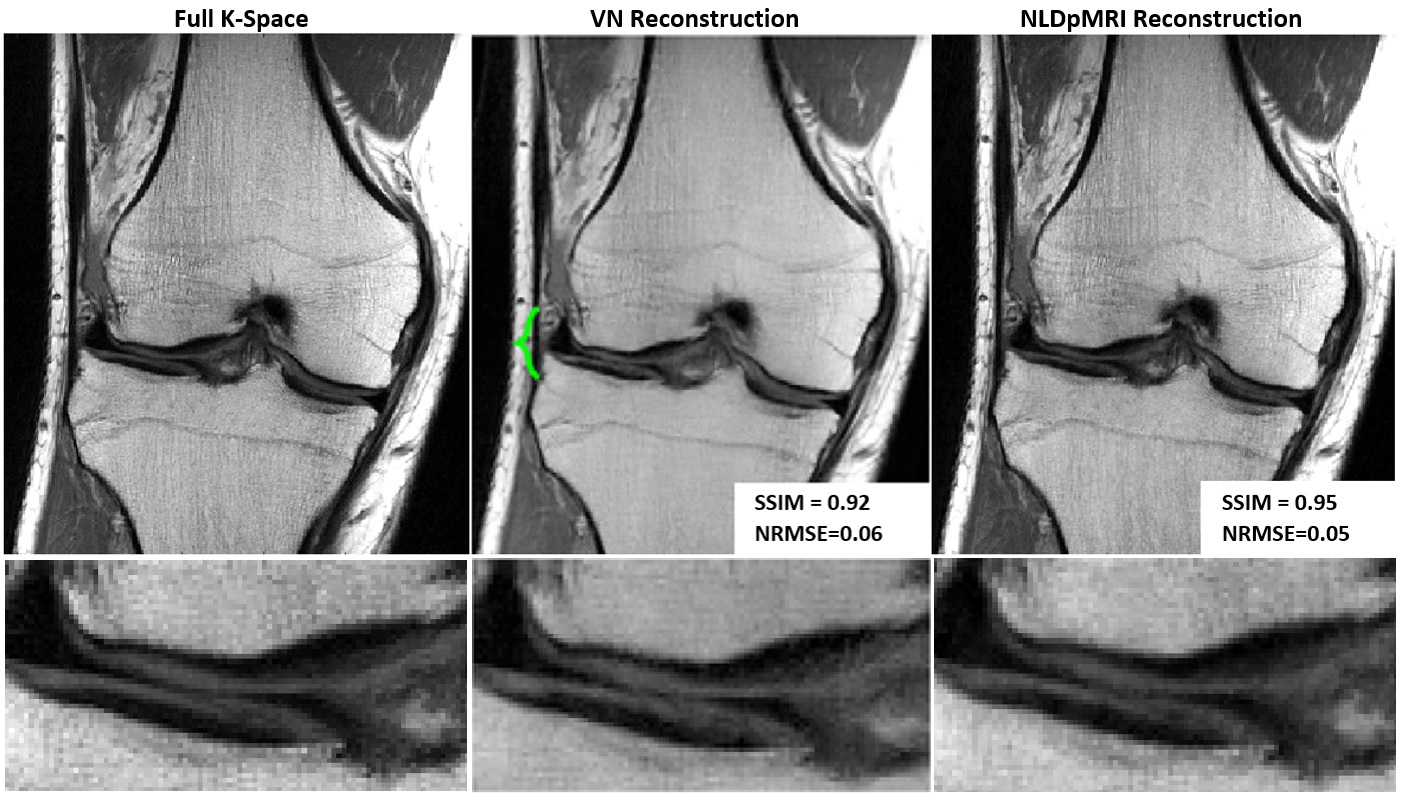}
\end{tabular}
\end{center}
\caption[fig4] 
{ \label{fig4} 
Left to right: Gold standard reconstruction result using full k-space data, VN reconstruction result, and regularized NLDpMRI reconstruction result all with undersampling factor of 4 for one case from coronal PD-weighted dataset. The green bracket indicates osteoarthritis. The second row shows the zoomed views. NLDpMRI results in slightly sharper image compared to VN.}
\end{figure}

\begin{figure} [ht]
\begin{center}
\begin{tabular}{c} %% tabular useful for creating an array of images 
\includegraphics[height=9cm]{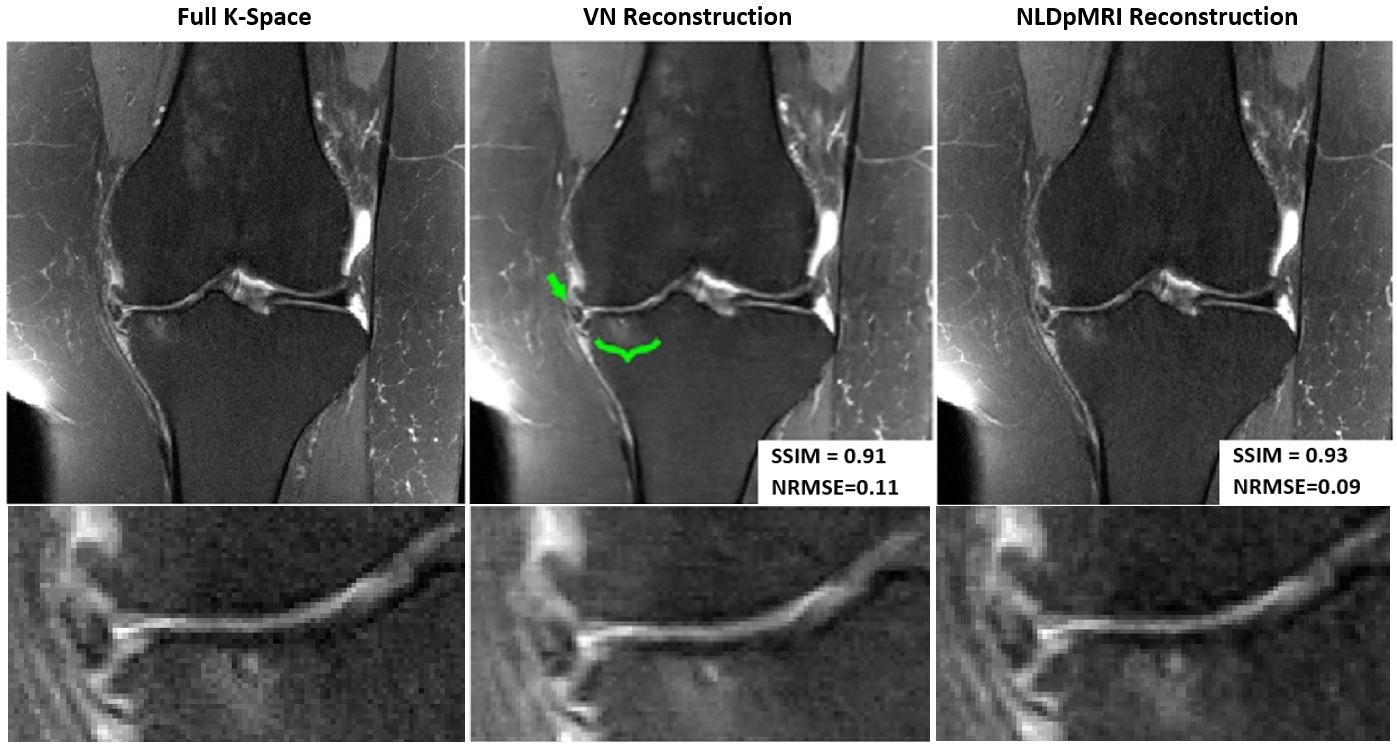}
\end{tabular}
\end{center}
\caption[fig5] 
{ \label{fig5} 
Left to right: Gold standard reconstruction result using full k-space data, VN reconstruction result, and regularized NLDpMRI reconstruction result all with undersampling factor of 4 for one case from coronal FS PD-weighted dataset. The green bracket shows broad-based, full-thickness chondral loss and a subchondral cystic change. The green arrow shows an extruded and torn medial meniscus. The second row shows the zoomed views. NLDpMRI results in slightly sharper image compared to VN.}
\end{figure}

Figure \ref{fig4} shows the impact of the acceleration factor of 4 for VN method, and NLDpMRI method on coronal PD-weighted data. The NLDpMRI result in Figure \ref{fig4} outperforms the learned VN result and generate sharper and higher quality reconstruction. A similar observation can be made for coronal FS PD-weighted data in Figure \ref{fig5}. The same sensitivities estimated from exactly the same calibration data have been used for both reconstruction methods in this experiment. Table \ref{table1} provides the quantitative evaluation for all five knee datasets using NLDpMRI and VN reconstructions. The NLDpMRI reconstruction show superior performance in terms of SSIM and NRMSE for four datasets (out of five datasets). The VN method shows slightly better performance compared to our method for saggital PD dataset.
Considering the fact that no training datasets have been used for NLDpMRI reconstructions in different experiments including different sequences with different parameters, proves the capability of NLDpMRI as a generalized parallel imaging method.

\section{CONCLUSIONS}

We propose a generalized method to solve MRI parallel image reconstruction problem using deep neural networks without any training data involved. The proposed approach eliminates the need to collect massive datasets for training purposes, any form of normalization, and transfer learning techniques to bring the data into the same domain as the trained network. Experimental results on real MRI acquisitions show that our proposed method outperforms the clinical gold standard GRAPPA method and the deep learning-based VN method.

\begin{table}[h!]
\centering
\begin{tabular}{rllll}
\hline
\multicolumn{3}{c} {\ NLDpMRI} & {VN} \\
\cline{2-3} 
\cline{4-5} 
Dataset & SSIM(\%) & NRMSE  & SSIM(\%) & NRMSE   \\
\hline
Coronal PD & $\textbf{94.76}\pm\textbf{1.38}$ & $\textbf{0.04}\pm\textbf{0.005}$  & $92.14\pm1.68$ & $0.08\pm0.02$   \\
Coronal FS PD & $\textbf{90.40}\pm\textbf{2.56}$ & $\textbf{0.10}\pm\textbf{0.01}$  & $81.97\pm3.60$ & $0.17\pm0.03$  \\
Sagittal FS T2 & $\textbf{94.55}\pm\textbf{0.89}$ & $\textbf{0.09}\pm\textbf{0.01}$  & $92.83\pm2.40$ & $0.11\pm0.03$  \\
Axial FS T2 & $\textbf{88.79}\pm\textbf{2.01}$ & $\textbf{0.16}\pm\textbf{0.02}$  & $88.46\pm2.43$ & $\textbf{0.16}\pm\textbf{0.02}$  \\
Sagittal PD & $94.33\pm0.95$ & $0.05\pm0.005$  & $\textbf{96.64}\pm\textbf{1.16}$ & $\textbf{0.04}\pm\textbf{0.01}$ \\

\hline
\end{tabular}
\caption{Quantitative evaluation on five different knee datasets in terms of SSIM and NRMSE using NLDpMRI and VN reconstruction methods.}
\label{table1}
\end{table}

\acknowledgments % equivalent to \section*{ACKNOWLEDGMENTS}       
 
This research was supported in part by NIH grants R01 NS079788, R01 EB019483, R42 MH086984, and by a research grant from the Boston Children's Hospital Translational Research Program.  

% References
\bibliography{report} % bibliography data in report.bib
\bibliographystyle{spiebib} % makes bibtex use spiebib.bst

\end{document}